\documentclass[fullpaper]{nldl}

\renewcommand{\DOI}{12345}

\usepackage[utf8]{inputenc}
\usepackage{url}
\usepackage{graphicx}
\usepackage{authblk}

\usepackage{array}
\usepackage{algorithm}
\usepackage{algorithmic}

\usepackage[square,numbers]{natbib}

\usepackage{hyperref}

\title{Lightweight Face Recognition: An Improved MobileFaceNet Model}
\author[1]{Ahmad Hassanpour}
\author[2]{Ahmad Hassanpour, Yasamin Kowsari}

\date{\vspace{-5ex}}

\begin{document}
\nldlmaketitle

\begin{abstract}  
This paper presents an extensive exploration and comparative analysis of lightweight face recognition (FR) models, specifically focusing on MobileFaceNet and its modified variant, MMobileFaceNet. The need for efficient FR models on devices with limited computational resources has led to the development of models with reduced memory footprints and computational demands without sacrificing accuracy. Our research delves into the impact of dataset selection, model architecture, and optimization algorithms on the performance of FR models. We highlight our participation in the EFaR-2023 competition, where our models showcased exceptional performance, particularly in categories restricted by the number of parameters. By employing a subset of the Webface42M dataset and integrating sharpness-aware minimization (SAM) optimization, we achieved significant improvements in accuracy across various benchmarks, including those that test for cross-pose, cross-age, and cross-ethnicity performance. The results underscore the efficacy of our approach in crafting models that are not only computationally efficient but also maintain high accuracy in diverse conditions.
\end{abstract}

\section{Introduction}
Efficient deep learning models are designed to optimize computational resources, reduce energy consumption, and maintain high levels of accuracy, thereby enabling advanced AI applications on resource-constrained devices such as smartphones and embedded systems \cite{menghani2023efficient, hassanpour2019novel, hasanpour2020software, rahimi2023toward, kowsari2020classification, daryani2023irl, hassanpourimpact, hassanpour2022differential}. Face recognition (FR) is widely preferred as a biometric recognition technique due to its non-intrusive nature and the exceptional accuracy achieved. Face recognition models have been extensively integrated into diverse domains such as automated border control, surveillance, and convenient application. Indeed, the performance of State-Of-The-Art (SOTA) face recognition models has witnessed significant improvements, even in challenging conditions \cite{kolf2023efar, hassanpoursynthetic}. The advancements in face recognition accuracy achieved by the latest  FR models can be attributed to the progress in deep learning network structures, training losses, and the accessibility of substantial identity-labeled training datasets. These factors have contributed significantly to the evolution of FR technology \cite{zhao2003face, hassanpour2022e2f, hassanpour4578828e2f}. Indeed, despite the progress made in the field of face recognition and the development of extremely large models, deploying such models on embedded devices and other use cases with limited computational capabilities and high throughput requirements continues the challenges. The memory requirements of these large models can be a significant constraint in such scenarios \cite{ben2021efficient}. In order to employ Deep Neural Networks (DNN) on embedded devices, specific requirements must be fulfilled regarding both the computational complexity, evaluated by the count of floating-point operations (FLOPs), and the memory usage measured in megabytes (MB) \cite{yan2019vargfacenet}.  Several efficient FR models have been proposed, and the core idea of many of these works depended on utilizing efficient architecture, which reduces the computational effort compared to larger and more complex DNNs.

Lately, there was a competition focused on efficient face recognition models \cite{kolf2023efar}. They divided the competition into two categories based on how big the models could be: one for models with less than two million parameters and another for models with 2-5 million parameters. We took part in this com- petition, and in certain tests, our models performed exceptionally well, securing top positions. In this paper, we aim to provide a more comprehensive exploration of the details behind our solutions and share a detailed account of our achievements during our participation in this competition.

\section{Related Works}
In the landscape of face recognition (FR) technology, the pursuit of models that balance efficiency with high accuracy has led to a proliferation of research focused on convolutional neural network (CNN) architectures. One significant stride in this domain is the reduction of model complexity through the elimination of fully connected (FC) layers at the top of CNNs, an approach that mitigates computational cost without substantially compromising performance \cite{li2019airface}. Recent literature has also seen the advent of MobileNets, which leverage depth-wise separable convolutions to streamline the network structure, and VarGFaceNet, which builds upon the VarGNet by introducing variable group convolutional networks to enhance efficiency. Notably, MobileFaceNets have garnered attention for achieving an impressive 99.55\% accuracy on the LFW dataset with less than 1M parameters, making them suitable for real-time applications. The augmentation of these architectures with residual bottlenecks and depth-wise separable convolutions has further refined the trade-off between the number of parameters and computational expense \cite{duong2019mobiface}.

Parallel to these developments, innovative frameworks like MixFaceNets \cite{boutros2021mixfacenets} have emerged, extending the MixConv block with a channel shuffle operation to increase discriminatory power, while maintaining a low parameter count. LightCNN's exploration of various architectural depths, ranging from 4 to 29 layers, demonstrated that deeper configurations could achieve better accuracy on benchmarks like the LFW dataset, albeit with an increased number of parameters. These models showcase the industry's ongoing efforts to strike a delicate balance between precision and efficiency, a core concern given the growing demand for deploying FR technologies in embedded systems and other resource-constrained environments. The body of work underpinning this paper's research draws on these pivotal studies, setting the stage for our own contributions to the field of efficient FR models.

\section{Our Solution}
In EFaR-2023 competition, the primary criterion was to have a model with a limited number of parameters. With this in mind, we adopted the MobileFaceNet model, boasting 1.1M parameters. Our pivotal contributions revolved around selecting the right dataset, picking a suitable deep learning model, optimizing algorithms, and refining the training procedure. We delve into each of these aspects in the subsequent subsections.
\subsection{Dataset}
Concerning the dataset, we opted for a subset of Webface42M. From this, we made a random selection of 100K individual identities, with each identity containing between 30 to 50 images. This choice ensured a rich variety of data, allowing our model to train on diverse samples, thereby enhancing its generalization capabilities. Notably, this subset encompassed a staggering 4.2M images from the broader WebFace42M dataset, further emphasizing the vastness and diversity of our selected data pool.
\subsection{Model Architecture}
For the initial category of the competition, which required models with fewer than 2M parameters, we employed the unaltered MobileFaceNet architecture. However, for the second category, we amplified the kernel count in all CNN layers by a factor of two. Despite these modifications, we retained the embedding feature vector from the original MobileFaceNet, which stands at 512. As a result of these changes, the total parameter count rose to 2.1M. We have named this enhanced architecture Modified-MobileFaceNet.
\subsection{Optimization Algorithm}
At this juncture, rather than relying on the conventional stochastic gradient descent for the optimization procedure, we employed the sharpness-aware minimization (SAM) optimizer for both models, setting $\rho$ at 0.02. Additionally, during the training phase, we incorporated an exponential learning rate scheduler with a gamma value of 0.998. SAM is a unique optimizer that focuses on enhancing generalization by reducing the sharpness of the loss landscape. Sharp minima in the loss landscape often lead to poor generalization on out-of-sample data. By flattening these sharp regions, SAM promotes a model that is more robust and generalizes better to unseen data.
\subsection{training process}
For both categories ($<$2M and 2-5M), we paused the training process on two separate epochs, first at epoch 20 and then again at epoch 50, even though the training continued for a total of 100 epochs. During these pauses, the learning rate was adjusted downward, first from 0.1 to 0.01 and subsequently to 0.001. Upon concluding the training, we employed datasets like LFW, CALFW, and AgeDB30 to discern and select the two best-performing models.
\section{Results}

This section presents and contrasts the outcomes of the baselines with the suggested solutions across various benchmarks. Subsequently, this review is extended with evaluations on bias, cross-quality, and a comprehensive recognition benchmark.

since we use a subset of WebFace42M, we will compared our results with the models (MobileFaceNet) trained on MS1MV2 dataset \cite{deng2019arcface} and consider this as baseline. Moreover, we will compared our results when the used optimization algorithm is SAM or SGD. Table 2 provides an overview of all the models, highlighting their variations. It details the submitted approaches, encompassing the core model design, the loss functions applied, the chosen optimizers, the datasets utilized for training, and the feature dimensions employed.
\begin{figure}
    \centering
    \includegraphics[width=0.9\linewidth]{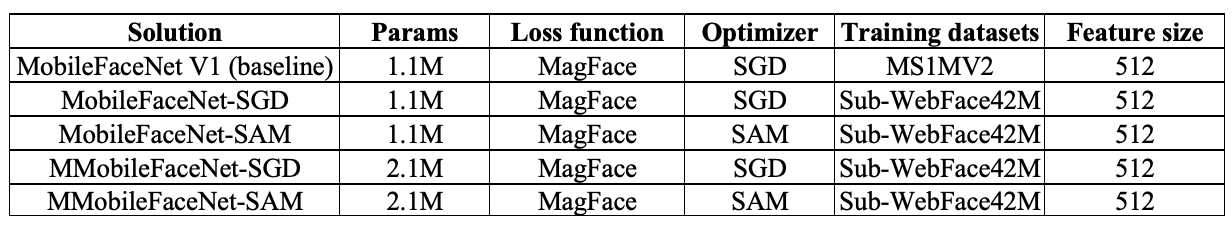}
    \caption{Details of solutions including the used loss function, optimizer and datasets during training, and the feature size used. All solutions use MobileFaceNet as architecture and 112 × 112 image size.}
    \label{fig:enter-label}
\end{figure}
The evaluation of both the proposed models and the baselines is delineated in Table 3. This table not only showcases the performance metrics but also provides insights into the FLOPs, model dimensions, and the total number of parameters employed. 
A significant observation from the results is the enhanced performance of models optimized using the SAM algorithm. On an average, SAM-optimized models with 1.1MP surpassed their counterparts by achieving approximately 4\% greater accuracy when compared to the baseline models that were trained on the MS1MV2 dataset. Furthermore, when juxtaposed with SGD-based models, the SAM-optimized models consistently demonstrated superior results. They surpassed the SGD-based models by an average accuracy margin of 1.2\%. 
This trend of SAM's superiority remains consistent even for models with 2.1MP. In this category, SGD-based models lagged behind by roughly 1.4\% in terms of accuracy when compared to their SAM-based counterparts. The consistent performance advantage of SAM-optimized models across different configurations underscores their robustness and efficacy in the given context.   
\begin{figure}
    \centering
    \includegraphics[width=0.9\linewidth]{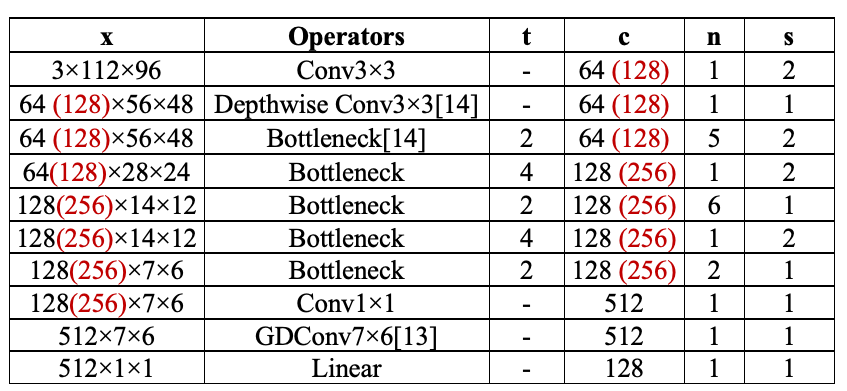}
    \caption{Architecture of MobileFaceNet V1 and modified (MMobileFaceNet) in channel numbers highlighted in red. }
    \label{fig:enter-label}
\end{figure}

\begin{figure*}
    \centering
    \includegraphics[width=1\linewidth]{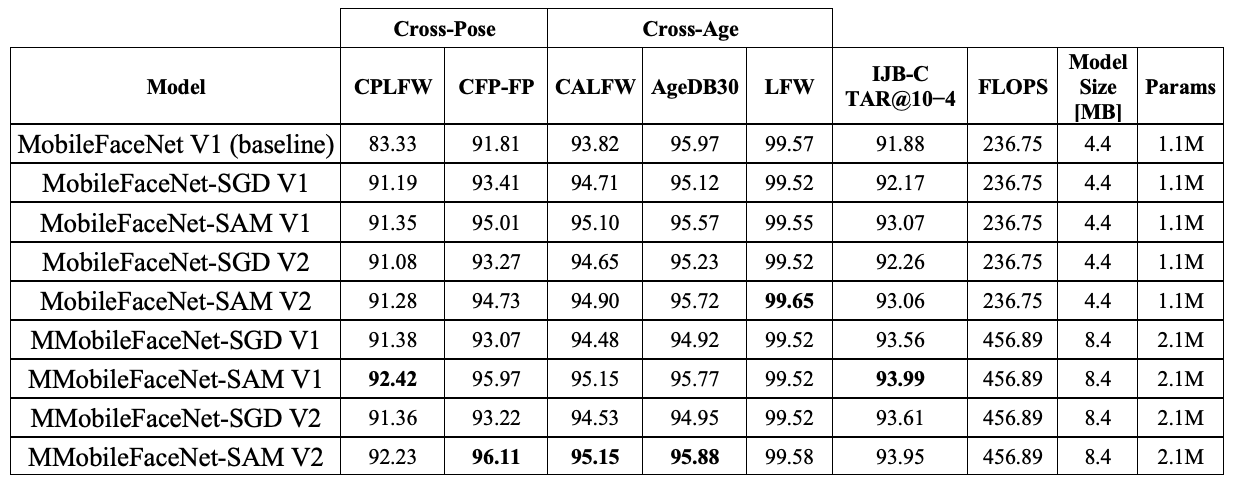}
    \caption{presents the evaluation results of our solutions in comparison to the baseline on benchmark datasets, including FLOPs, model size, and the number of parameters. The best accuracies for each dataset are highlighted.}
    \label{fig:enter-label}
\end{figure*}

Table 4 presents the assessment outcomes of the provided solutions across the RFW, XQLFW, and IJB-C benchmarks. Within the RFW metric, verification accuracies are specified for four ethnic subgroups: Asian, African, Caucasian, and Indian, from which an average accuracy and its standard deviation are derived. The SER metric highlights a model's bias, with a higher value suggesting a pronounced bias. For the XQLFW benchmark, the emphasis is on verification accuracy. TinyFaces displays both rank 1 and rank 5 measurements, while IJB-C showcases the TAR at various FAR levels. Notably, baseline figures are incorporated for every dataset. In categorizing by parameter count, submissions with parameters ranging between 2M and 5M are denoted as "2-5 MP", while those with fewer than 2M parameters are labeled "2 MP".

In the context of the RFW dataset, it remains evident that models optimized with the SAM algorithm consistently surpass those founded on the SGD methodology. Specifically, when evaluating low-resolution images, a noticeable disparity of approximately 5\% emerges between the baseline and our advanced solutions. This gap is observed across both the SGD and SAM-optimized models, irrespective of whether they possess 1.1MP or 2.1MP. This underlines the enhanced capability of our proposed models, showcasing their superior performance in handling various image resolutions and model complexities.

Table 3 provides a detailed performance analysis of different iterations of MobileFaceNet and MMobileFaceNet models, highlighting the best accuracies in bold for easier identification. These accuracies are evaluated against a variety of benchmark datasets that test factors like cross-pose and cross-age variations, with datasets such as CPLFW, CFP-FP, CALFW, AgeDB30, LFW, and IJB-C. The best accuracies for each dataset are emboldened, indicating superior performance in those specific tests. For instance, MMobileFaceNet-SAM V1 and MMobileFaceNet-SGD V2 show prominent bold figures, suggesting they outperform other models in several categories. This layout allows for a quick assessment of which model versions achieve peak performance, providing insights into the effectiveness of the different models' architectures and training methods. The consistency in model size and parameters across the MobileFaceNet versions and the increased size and parameters for the MobileFaceNet variants are also notable, suggesting a trade-off between model complexity and accuracy.

Table 4 the results from evaluating various face recognition models, specifically focusing on the MobileFaceNet and MMobileFaceNet architectures, across different benchmarks that test their accuracy on low-resolution images and across different ethnic groups. The benchmarks used here are RFW (Racial Faces in the Wild), XQLFW (Cross Quality Labeled Faces in the Wild), and IJB-C (IARPA Janus Benchmark-C). For the RFW benchmark, which tests verification accuracy across four ethnic subgroups (Asian, African, Caucasian, and Indian), the table lists the accuracies for each model within each subgroup, along with the average accuracy across all groups. The accuracies for the best-performing models within each subgroup are highlighted in bold, indicating the highest accuracy achieved for that particular ethnicity. For instance, MMobileFaceNet-SGD V2 and MMobileFaceNet-SAM V1 show high accuracy across most ethnicities, with their performance in the Asian and African categories being notably superior. In the low-resolution XQLFW benchmark, the models are evaluated on their verification accuracy, and the table shows two metrics: the rank-1 accuracy percentage and the True Accept Rate (TAR) at different False Accept Rates (FAR), specifically FAR=$10^-5$, FAR=$10^-4$, FAR=$10^-3$, and FAR=$10^-2$. The highest accuracies are again highlighted, showcasing which models perform best at matching low-resolution images. The IJB-C benchmark results are presented in the form of TAR at various FARs. This benchmark is particularly challenging as it involves a mix of image and video data, representing real-world scenarios that require robust face recognition capabilities. The MMobileFaceNet models seem to excel in this benchmark as well, achieving high accuracy rates, as indicated by the bold numbers, especially at lower FARs.
\begin{figure*}
    \centering
    \includegraphics[width=1\linewidth]{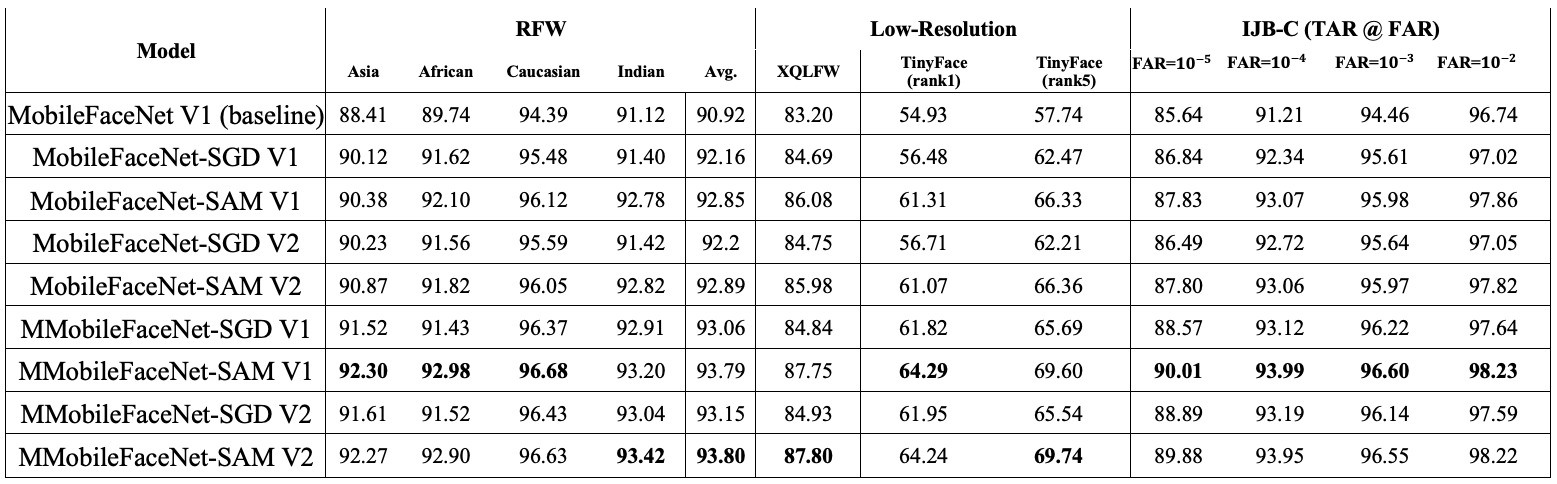}
    \caption{We conducted evaluations of our solutions and the baseline performance on the RFW, XQLFW, and IJB-C benchmarks. In the case of RFW, verification accuracy is provided for four distinct ethnic subgroups: Asian, African, Caucasian, and Indian. The average accuracy and standard deviation were calculated based on the performance across these four ethnicity datasets. Meanwhile, for XQLFW, we report the verification accuracy.}
    \label{fig:enter-label}
\end{figure*}

\section{Conclusion}
In conclusion, our study has successfully demonstrated that it is possible to design and train highly accurate and computationally efficient FR models. Our modified MobileFaceNet architectures, optimized with SAM, consistently outperformed baseline models across a suite of challenging benchmarks. The evaluation results, particularly in the RFW and XQLFW benchmarks, have shown the robustness of our models against variations in pose, age, and ethnicity, as well as their ability to handle low-resolution images effectively. Furthermore, the IJB-C benchmark results highlighted the models' excellent true accept rates at various false accept rates, showcasing their potential for real-world applications. These findings affirm the potential of fine-tuned lightweight models in settings where computational resources are constrained, paving the way for broader deployment of FR technology across a variety of platforms.

\bibliographystyle{abbrvnat}Architecture Loss function Optimizer Training datasets Feature size
\bibliography{references}

\end{document}